\documentclass[sigconf]{acmart}

\AtBeginDocument{%
  }

\setcopyright{acmlicensed}
\copyrightyear{2026}
\acmYear{2026}
\acmDOI{XXXXXXX.XXXXXXX}

\acmConference[CHI '26]{CHI Conference on Human Factors in Computing Systems}{April 13--17, 2026}{Barcelona, Spain}
\acmBooktitle{Proceedings of the CHI 2026 Workshop: Ethics at the Front-End: Responsible User-Facing Design for AI Systems (CHI '26), April 13--17, 2026, Barcelona, Spain}
\acmISBN{978-1-4503-XXXX-X/2026/04}

\begin{document}

\title{Designing for Disagreement: Front-End Guardrails for Assistance Allocation in LLM-Enabled Robots}

\author{Carmen Ng}
\email{carmen.ng@tum.de}
\affiliation{%
  \institution{Technical University of Munich}
  \country{Germany}
}

\begin{abstract}
LLM-enabled robots prioritizing scarce assistance in social settings face pluralistic values and LLM behavioral variability: reasonable people can disagree about who is helped first, while LLM-mediated interaction policies vary across prompts, contexts, and groups in ways that are difficult to anticipate or verify at contact point. Yet user-facing guardrails for real-time, multi-user assistance allocation remain under-specified. We propose \emph{bounded calibration with contestability}, a procedural front-end pattern that (i)~constrains prioritization to a governance-approved menu of admissible modes, (ii)~keeps the active mode legible in interaction-relevant terms at the point of deferral, and (iii)~provides an outcome-specific contest pathway without renegotiating the global rule. Treating pluralism and LLM uncertainty as standing conditions, the pattern avoids both silent defaults that hide implicit value skews and wide-open user-configurable ``value settings'' that shift burden under time pressure. We illustrate the pattern with a public-concourse robot vignette and outline an evaluation agenda centered on legibility, procedural legitimacy, and actionability, including risks of automation bias and uneven usability of contest channels.
\end{abstract}



\maketitle

\section{Introduction}

As large language models (LLMs) are increasingly embedded into socially assistive robots as components shaping high-level decision-making, commonsense reasoning, and action-selection~\cite{ahn2022,jeong2024,wang2025llmrobot,zhang2023}, a front-end ethics challenge becomes harder to treat as a back-end-only issue: LLM behavior no longer shapes only static outputs but also \textbf{interaction policy} in open-world settings, extending engagement and task sequencing toward determining who is acknowledged, deferred, and assisted first, thereby in effect allocating scarce attention or assistance in real time. Social robots are already deployed across care, service, and public navigation domains~\cite{ahmed2024,lambert2020,lee2021service,savela2018}, including recurrent multi-party settings~\cite{garcia-haro2020,sorrentino2024}. In edge cases with competing needs and limited time, these sequencing decisions can affect access to help and perceived fairness across diverse social norms, often without legible rules or usable avenues for contestation. While concentrated in robotics research and emerging commercial systems via multimodal stacks~\cite{elish2019,gemini2025}, the expanding LLM-robot convergence introduces model variability and stochasticity into embodied interaction contexts, amid emerging risk signals at both model and interface levels across fields: audits of LLM-driven robots flag group-based discrimination under open-vocabulary inputs~\cite{hundt2025}; LLM studies uncover social bias~\cite{gallegos2024} and uneven value generalization across populations and languages~\cite{adilazuarda2024,cao2023,fraser2022}; HCI research shows interaction design can encode harms via manipulation and exclusion~\cite{gray2018}. While model-level mitigations remain crucial, this paper focuses on \textbf{ethical safeguards operationalized through front-end mechanisms} supporting transparency, user agency, and contestability, e.g., what users can see, understand, and act upon at the point of contact (or deferral) with an LLM-enabled social robot~\cite{abdul2018}, rather than assigning ethical responsibility to model properties alone. We introduce \textbf{bounded calibration with contestability} as a front-end design pattern for assistance allocation, featuring a governance-approved menu of prioritization modes, legibility throughout interaction, and a contestation pathway, preventing silent defaults while making value-laden sequencing inspectable and procedurally accountable.

\section{Related Work}

Adjacent literature provides building blocks for ethical front-end design, but they generally stop short of mechanism-level guidance for \emph{interaction-time assistance allocation} when an LLM-enabled robot must sequence help under scarcity and situational uncertainty. The gap is not ethical intent, but rather an under-specification of how an embodiment system's front end should make a prioritization rule legible, keep it within admissible bounds, and provide usable challenge pathways when the allocation is enacted through deferral in the moment. HCI and human-centered AI work shows that front-end configuration is not neutral: small interface choices can shift outcomes in consent and choice architectures~\cite{mathur2019,nouwens2020}. By design rationale, allocation policy in LLM-enabled robots is similarly implemented as ``small'' interaction moves (who gets acknowledged, and how this is justified), so even silent defaults already function as value-laden governance choices rather than mere technical parameters. Studies also suggest transparency is experienced through procedural features rather than a binary ``disclosed vs.\ not disclosed'' property~\cite{rader2018}, implying that legibility of a prioritization mode is a front-end design problem. System-level syntheses further argue that accountable systems require user-facing interaction mechanisms, not merely algorithmic techniques~\cite{abdul2018}. Yet systematic mapping shows that responsible AI work clusters around high-level governance~\cite{tahaei2023}, offering limited guidance for \textbf{interaction-level guardrails} as LLM integration can shift robot interaction patterns from predefined rules toward context-sensitive, language-mediated reasoning~\cite{kim2024,lee2023empathy}.

In parallel, multi-user HRI studies show that interaction policy is designable in shared-robot settings, such as using engagement and turn-taking policies to determine who is addressed and when, and conflict handling can shape user evaluations~\cite{moujahid2022,soderlund2024}. However, these policies are more often treated as coordination or social intelligence problems than as \emph{distributive} commitments that should be explicitly governed as a front-end ethical interface. Prior work on procedural justice and contestability emphasizes that fairness and legitimacy depend on process features and usable procedures, not only outcomes or formal appeal rights alone~\cite{lee2019procedural,lyons2021,yurrita2025}. Yet much of this guidance is developed around non-embodied (e.g., online platforms) or post-hoc decision settings, leaving open questions on contestability in interaction-time deferral with material stakes. We clarify a scope boundary: our argument does not depend on how contention is detected (overlapping speech or sensor inference). Our claim is narrower: when contention occurs, prioritization is operationalized through interaction; under LLM behavioral uncertainty, governance must be available through front-end legibility and recourse. In sum, existing literature provides components including value-laden interface mechanisms, multi-user interaction policy, and procedural legitimacy, but they leave under-specified an integrated front-end mechanism anchored to real-time assistance allocation uniquely relevant for LLM-enabled robots and their diverse users.

\section{Bounded Calibration With Contestability}

\subsection{Centering pluralism and uncertainty}

Under scarcity, an embodied agent inevitably allocates limited attention through interaction. Because reasonable prioritization principles frequently conflict (e.g., urgency-first, queue order, vulnerable groups-first), any silent default becomes a non-neutral value commitment. This matters because value pluralism is a standing condition, not an edge case. Fairness judgments vary within populations and across contexts, shaped by outcome favorability and individual differences~\cite{wang2020fairness} and rarely converging on a single interpretation~\cite{starke2022}. Cross-cultural work similarly cautions against assuming universality in how transparency or fairness are interpreted~\cite{chien2025}. Meanwhile, multilingual LLM studies report cross-cultural biases and value misalignment~\cite{rystrom2025,tao2024}, so ``cultural competence'' is not a safe default to outsource to LLM behavior. Accordingly, we do not claim a universally correct rule; we instead treat \textbf{value pluralism and LLM behavioral uncertainty as conditions the front end must govern}. Under these conditions, leaving any single rule as a silent default would conceal value commitments, while full user configurability would invite coercion, preference conflicts, and burden-shifting towards users under time pressure. The ethical front-end alternative is therefore user-facing, bounded governance.

\subsection{Pattern overview: what bounded calibration means}

We propose \textbf{bounded calibration with contestability} as a front-end pattern coupling three elements: a governance-approved set of prioritization modes, continuous mode legibility in interaction-relevant terms, and a lightweight pathway to challenge or escalate outcomes. Although prioritization is executed by back-end components, we treat it as a front-end ethics problem here since legitimacy and perceived fairness depend on process features, not outcomes alone~\cite{lee2019procedural}. Importantly, we \textbf{separate value mediation from interaction-time allocation}, as real-time ``value balancing'' can reintroduce opaque trade-offs when fairness depends on abstraction and context choices~\cite{selbst2019}. We therefore structure value mediation across three governance layers: (i)~\textbf{Define}: deployers define a small set of defensible prioritization modes and exclude harmful configurations as upstream boundaries, reflecting critiques that high-level principles often underdetermine implementable rules in practice~\cite{mittelstadt2019}; (ii)~\textbf{Select}: authorized roles choose the active mode for a context window (e.g., time shift, location), with role-gating and rate limits to preserve predictability and avoid preference conflicts; (iii)~\textbf{Challenge}: users can contest a specific deferral and demand escalation without re-negotiating the global rule, consistent with findings that meaningful contestation requires concrete, usable mechanisms, and transparent review procedures~\cite{lyons2021,yurrita2025}.

\textbf{Bounded} means calibration is constrained along three dimensions: (i)~\textbf{Admissibility} (which modes can be chosen): calibration is not an open ``values settings'' panel; it restricts prioritization to institutionally sanctioned modes and excludes extreme or discriminatory configurations. This responds to evidence that choice architectures can be engineered to steer or obstruct decisions through dark patterns~\cite{mathur2019,nouwens2020}; (ii)~\textbf{Abstraction} (what level is chosen): calibration does not operate as step-by-step micro-control, but at the level of prioritization principles (e.g., urgency-first vs queue-order), avoiding overly-granular fairness rules that can be context-blind in diverse societal environments~\cite{selbst2019}; (iii)~\textbf{Authority and timing} (who can change it and when): calibration is governance-constrained rather than individually configurable; mode switching is role-gated and rate-limited, while user voice is integrated through contestability, not instant overrides. This also guards against responsibility displacement where humans afforded limited control can become the blamed surface~\cite{elish2019}.

\section{Scenario Vignette: LLM-Enabled Robot Guide in a Busy Concourse}

This illustrative scenario shows how bounded calibration, mode legibility, and contestability can jointly govern scarcity-driven allocation at the front end (abstracting over input modality):

\textbf{Setup (bounded mode selection):} A public guide robot in a busy concourse connecting a train station to a mall can attend to only one interaction at a time at peak hours, creating contention. The station therefore pre-defines an admissible menu of prioritization modes and authorizes staff to select one for a time window. At the start of the shift, staff select an ``urgency-first'' mode from this bounded, policy-approved menu (e.g., urgency-first, queue-order, vulnerability-aware), allowing legitimate variation across contexts without making any principle a silent default.

\textbf{Allocation point (legibility at deferral):} Two requests arrive in close succession: a tourist asks for directions, another distressed person reports a lost wallet. The robot prioritizes the distressed person and defers the tourist while disclosing the active mode (``\emph{Priority mode: urgent needs first --- I'll return to you next}''), aligning with HRI transparency and explainability work that frames understanding as a communicative design problem in co-located settings~\cite{schott2023}.

\textbf{Contest point (outcome-specific recourse):} The tourist contests the deferral (e.g., via a spoken phrase, a button, or an operator channel). Contestation does not necessarily change the global mode, but instead triggers an outcome-specific pathway communicating the grounds and consequences of challenge, e.g., a brief clarification and optional escalation to staff, aligned with research on how meaningful challenge demands usable mechanisms, not only appeal rights~\cite{lyons2021,yurrita2025}.

\textbf{Boundaries and trace to enable stability and reviewability:} If the tourist attempts to switch modes, the robot role-gates the action (``\emph{Only station staff can change priority mode}''); discriminatory or inadmissible modes are rejected by default. The interaction is also logged (active mode, deferral, contestation, escalation outcome) to support later review, echoing procedural fairness work that emphasizes reviewable processes instead of outcomes alone~\cite{lee2019procedural}.

\section{Evaluation Agenda}

This paper does not report an empirical evaluation; instead, we outline evaluation targets focusing on three benchmarks: \textbf{legibility} (mode comprehension: can users identify the active mode and anticipate deferrals), \textbf{legitimacy} (procedural fairness: do users judge allocation based on process features, not only on ``who gets help first''), and \textbf{actionability} (contestation: can decision subjects access and complete contest steps under time pressure, and understand what happens next). Feasible, diagnostic methods can isolate interface governance from back-end LLM capability, such as vignette experiments that compare silent defaults, legible prioritization modes, and legible modes plus contestability; Wizard-of-Oz multi-user studies that stress-test timing and interruption; or governance workshops that probe the feasibility of an admissible mode menu. Finally, because contest channels may be under-used if they are perceived as inefficient or futile, evaluation should also test adoption and drop-off across user groups and accessibility constraints, and whether contest logs and reviews actually feed back into organizational learning and future revision of admissible modes.

\section{Limitations}

Because prioritization principles remain contested even within a single deployment community, our contribution is procedural and targeting allocation settings faced by LLM-enabled robots in dynamic environments. We do not propose an empirically validated interface, nor any model-level alignment method. The pattern depends on governance capacity for mode definition and role-gating. We also acknowledge that contestability may be unevenly usable across user groups and access needs, and legible modes may induce automation bias over time. Finally, we scope the pattern to scarcity-driven allocation primarily relevant for LLM-enabled robots deployed in socially assistive settings, not for all AI systems.

\section{Conclusion and Future Pathways}

LLM-enabled robots enact assistance prioritization through interaction, rendering it an interface-level governance issue rather than a back-end challenge alone. Our contribution is a procedural front-end guardrail that constrains allocation to a governance-approved set of value choices, supports explainability during interaction, and provides a low-barrier path to contest and escalate without immediately re-negotiating the global rule. This pattern makes interaction policy inspectable and accountable where differential treatment is \emph{experienced}. \textbf{For deployers}, it offers a way to bound and disclose prioritization choices without exposing harmful configurations; \textbf{for researchers}, it defines evaluable front-end criteria beyond back-end performance; \textbf{for regulators and auditors}, it provides interaction-level traces of where and how value commitments are enacted, challenged, or revised, making ethical risks inspectable as LLM-robot convergence expands.

\bibliographystyle{ACM-Reference-Format}
\bibliography{references}

\end{document}